\pdfoutput=1

\documentclass[11pt]{article}

\usepackage[]{acl}

\usepackage{times}
\usepackage{latexsym}
\usepackage{enumitem}
\usepackage{CJKutf8}

\usepackage[T1]{fontenc}

\usepackage[utf8]{inputenc}

\usepackage{microtype}

\usepackage{inconsolata}

\usepackage{latexsym}
\usepackage{graphicx}
\usepackage{array}
\usepackage{amsmath, bm}
\usepackage[linesnumbered,ruled,vlined]{algorithm2e}
\usepackage{algorithmic}
\usepackage{multirow}
\usepackage[T1]{fontenc}
\usepackage{longtable}
\usepackage{booktabs}
\usepackage{threeparttable}

\usepackage[utf8]{inputenc}

\usepackage{microtype}
\newcommand{\hide}[1]{}
\newcommand{\vpara}[1]{\vspace{0.05in}\noindent \textbf{#1 }}

\title{PCQPR: Proactive Conversational Question Planning with Reflection}

  \author{Shasha Guo\textsuperscript{1, 2}\thanks{       $\text{ }$ This work was done during an internship at SMU.\normalsize}, 
 Lizi Liao\textsuperscript{3}, Jing Zhang\textsuperscript{1, 2}, 
\textbf{Cuiping Li\textsuperscript{1, 2}}, \textbf{Hong Chen\textsuperscript{1, 2}} \\
  \textsuperscript{1}School of Information, Renmin University of China, Beijing, China\\
  \textsuperscript{2}Key Laboratory of Data Engineering and Knowledge Engineering of Ministry of Education\\
  \textsuperscript{3}Singapore Management University\\
  \{guoshashaxing, zhang-jing, licuiping, chong\}@ruc.edu.cn, lzliao@smu.edu.sg
  }

\begin{document}
\maketitle
\begin{abstract}

Conversational Question Generation (CQG) enhances the interactivity of conversational question-answering systems in fields such as education, customer service, and entertainment. 
However, traditional CQG, focusing primarily on the immediate context, lacks the conversational foresight necessary to guide conversations toward specified conclusions. This limitation significantly restricts their ability to achieve conclusion-oriented conversational outcomes.
In this work, we redefine the CQG task as Conclusion-driven Conversational Question Generation (CCQG) by focusing on proactivity, not merely reacting to the unfolding conversation but actively steering it towards a conclusion-oriented question-answer pair. 
To address this, we propose a novel approach, called \textbf{P}roactive \textbf{C}onversational \textbf{Q}uestion \textbf{P}lanning with self-\textbf{R}efining (\textbf{PCQPR}). 
Concretely, by integrating a planning algorithm inspired by Monte Carlo Tree Search (MCTS) with the analytical capabilities of large language models (LLMs), PCQPR predicts future conversation turns and continuously refines its questioning strategies. This iterative self-refining mechanism ensures the generation of contextually relevant questions strategically devised to reach a specified outcome.
Our extensive evaluations demonstrate that PCQPR significantly surpasses existing CQG methods, marking a paradigm shift towards conclusion-oriented conversational question-answering systems.


\end{abstract}
\section{Introduction}
\label{sec:intro}
\begin{figure}[!t]
\centering 
\includegraphics[width=0.48\textwidth]{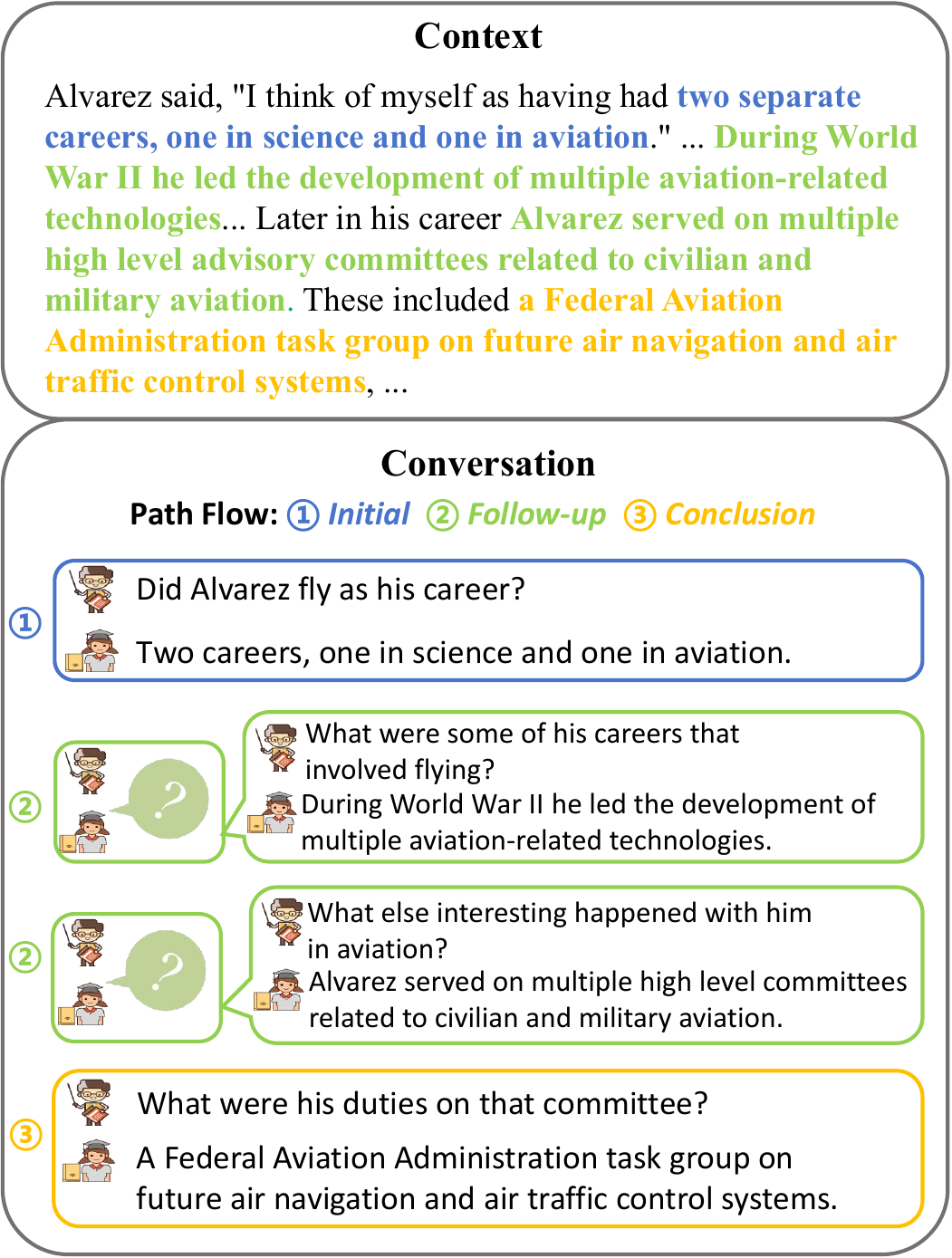}
\caption{An example of CCQG. Given the specific context, the initial conversation (blue), and the concluding question-answer pair (orange), CCQG proactively generates subsequent questions (green), aiming to advance toward the predefined question-answer pair.}
\label{fig:example} 
\end{figure}

Conversational Question Generation (CQG) has significantly enhanced the capabilities of conversational question-answering (QA) systems, bringing a level of dynamism and intelligence that was previously unattainable. In various fields such as educational technology, customer service, and interactive entertainment, CQG has become a crucial component for enhancing user interactions, enabling more natural and responsive conversations between humans and machines. By generating contextually appropriate questions, CQG systems have improved the interactive experience, making conversations with AI more engaging and informative.

Despite these advancements, current CQG methods are predominantly reactive, generating questions based solely on the immediate context without strategically considering the specified conversational outcome.
This reactivity is evident in various existing approaches: from choosing the most relevant conversational history to enhance question relevance~\cite{CQG}, to utilizing a context-enhanced neural model for identifying key contexts~\cite{CQGTOIS}, and designing a two-stage framework for generating more natural conversations~\cite{SG-CQG}.
While these methods contribute significantly to the field, their inherent reactivity limits their potential to direct conversations toward specific, predetermined conclusions.

These identified limitations underscore the necessity for a paradigm shift towards proactiveness in CQG, leading us to propose the novel task of Conclusion-driven Conversational Question Generation (CCQG), as illustrated in Figure~\ref{fig:example}. CCQG emphasizes the strategic generation of questions to guide conversations toward specific outcomes, a critical capability that reactive models fail to support adequately. By focusing on proactive question generation, we aim to overcome the limitations of reactivity and enable more purposeful and outcome-oriented question-answering conversations.

In contrast to conversational QA systems, the field of proactive dialogue systems has recently seen the emergence of several frameworks designed to accurately predict user needs and steer conversations accordingly.
Methods such as the application of MCTS for dialogue planning \cite{GDP-Zero, mcts2, mcts3}, and the use of large language models (LLMs) for predicting conversation trajectories \cite{ AskClarify,dao2023reinforced,liao2023proactive}, represent significant strides towards proactive dialogue management. However, these approaches do not directly address the nuanced requirements of CCQG, which relies on strategically generating questions to guide conversations toward specified outcomes.

To bridge this gap, we propose the \textbf{P}roactive \textbf{C}onversational \textbf{Q}uestion \textbf{P}lanning with self-\textbf{R}efining (\textbf{PCQPR}) framework, which innovatively integrates strategic planning and reflective refinement.
By incorporating an MCTS-like planning algorithm with the analytical capabilities of LLMs, our approach utilizes LLMs to predict and simulate future conversation paths, thus guiding the question generation toward the specified, desired outcome. 
Concurrently, LLMs engage in reflective analysis of these simulations, providing critical feedback for each action within the simulation paths to inform and enhance conversational strategies. 
This reflective analysis identifies failures and showcases successful experiences, providing a balanced and thorough evaluation of each planning path.
By adopting the comparable reflection mechanism that captures errors and successes, the framework guarantees the systematic incorporation of insights, leading to progressively enhanced question generation.
The PCQPR framework's innovative use of LLMs for both forward-looking simulation and reflective feedback marks a strategic shift from reactive to proactive CQG, aligning conversations more effectively with specified outcomes.

\vpara{Main Contributions.}(1) We propose a novel CCQG task that emphasizes proactive conversation steering in conversational question-answering systems. (2) We introduce an advanced PCQPR framework that uniquely integrates strategic planning with reflective refinement. Using MCTS-like algorithms and a reflection strategy, it predicts future conversation paths and provides critical feedback, continuously improving question generation. (3) Extensive experiments validate the effectiveness of PCQPR, demonstrating its superior performance and marking a notable advancement in proactive conversational question-answering systems.

\hide{
\begin{itemize}[nosep]
    \item We devise a novel CCQG task that emphasizes proactive conversation steering in conversational question-answering systems. 
    \item We propose the PCQPR framework, uniquely integrating strategic planning with reflective refinement. Using MCTS-like algorithms and a reflection strategy, it predicts future conversation paths and provides critical feedback, continuously improving question generation.
     \item Extensive experiments validate the effectiveness of PCQPR, demonstrating its superior performance and marking a notable advancement in proactive conversational QA systems.
\end{itemize}
}
\section{Related Work}
\label{sec:related}

\vpara{Question Generation.}
Question Generation (QG) aims to generate questions from diverse inputs, including text~\cite{ControllQG, DiversifyQG}, knowledge base~\cite{chen2020toward, PCQG, DSM, SGSH, Diversify_Guo}, and so on. 
Traditionally, QG has played a pivotal role in numerous practical applications, focused on producing contextually appropriate and meaningful questions~\cite{NQG_Survey}.

Recently, research has increasingly focused on Conversational Question Generation (CQG), which emphasizes multi-turn interactions to better simulate dynamic and natural conversational scenarios. In this domain, frameworks such as that proposed by ~\citet{ChainCQG} incorporate answer encoders and QG modules to learn from each conversational turn and generate subsequent questions. Similarly, ~\citet{CQG} apply top-$p$ strategies to select pertinent conversation history, thereby enhancing question relevance. Other notable approaches include ~\citet{CQGTOIS}'s context-enhanced neural model for identifying key contexts in question generation, and ~\citet{Zero-CQG}'s Zero-shot CQG, which utilizes transfer learning for multi-turn scenarios from single-turn QG instances. Furthermore, the `what to ask' and `how to ask' modules of ~\citet{SG-CQG}'s two-stage framework further highlight the evolving complexity in CQG.

Despite these developments, most CQG approaches remain reactive, focusing on generating questions from existing conversational content. To bridge this gap, we propose Conclusion-driven Conversational Question Generation (CCQG). CCQG departs from the reactive nature of traditional CQG, aiming to proactively generate a series of questions that guide the conversation toward a specified outcome. This novel task represents a strategic shift towards outcome-oriented conversation management in conversational QA systems.

The CCQG task holds significant practical value across various domains. For example, in intelligent tutoring systems, CCQG can guide students through structured sequences of questions, tailored to their current understanding, thereby personalizing and enhancing the educational experience. In Customer Service, particularly in troubleshooting scenarios, guiding a conversation toward a specific resolution is critical. Often, customers may not know the exact problem they face. A CCQG system can lead the customer through diagnostic questions, efficiently identifying the issue and guiding them to the appropriate solution. In interactive storytelling or role-playing games, the narrative's richness depends on how well the system can guide the user through the story. CCQG enhances user experience by maintaining narrative coherence and leading them to key plot points. By integrating CCQG into these systems, we aim to shift conversational interactions from passive question generation to an active, outcome-driven process.

\vpara{Planning and Reflection.} 
In the field of Natural Language Processing (NLP), 
planning has emerged as a crucial strategy, particularly for achieving specific targets through a series of intermediate actions. Recent advancements have seen the utilization of LLMs in various planning-based NLP tasks, evidenced by approaches such as Chain-of-Thought (CoT) reasoning~\cite{COT}, CoT with self-consistency~\cite{COT-SC}, Reasoning via Planning (RAP)~\cite{RAP}, and Tree of Thoughts (TOT)~\cite{TOT}. These methods illustrate the capability of LLMs to generate structured, step-by-step solutions. However, they often employ a linear, simplistic planning model that lacks iterative feedback, which can constrain their adaptability in dynamic NLP tasks.

To address the above challenges, recent approaches have incorporated reflection from LLMs as feedback~\cite{Self-Refine, shinn2023reflexion, he-etal-2024-planning} to refine planning processes. This shift represents a movement towards more dynamic planning, emphasizing the iterative improvement of individual plans. However, these methods primarily enhance single action sequences and do not fully explore multiple concurrent paths. In our work, PCQPR introduces a novel combination of a tree search algorithm with reflective refinement, facilitating complex, multi-path planning. This approach differs from traditional methods by evaluating various potential actions at each planning step, ensuring that the generated questions are precisely aligned with the desired outcome.



\section{Problem Formulation}
\label{sec:problem define}

A conversational question generation dataset\footnote{Since CCQG is a novel task, there is no directly available dataset. The original datasets do not explicitly state the predetermined conclusions, so we have restructured the data into the format we are currently using.} is denoted as $\mathcal{D} = \{C_i, H_i, T_i, \mathcal{G}_i\}_{i=1}^N$ with $N$ as the total number of conversations.
$C_i = \{s_1, s_2, ..., s_m\}$ with $m$ sentences represents the context related to the $i$-th conversation. 
$H_i = \{(q_1, an_1),(q_2, an_2), ...,(q_{t}, an_{t})\}$ with $t$ question-answer pairs represents the conversational history in the $i$-th conversation. 
$T_i$ denotes the predefined conclusion consisting of a question-answer pair $(q_{n}, an_{n})$. 
$\mathcal{G}_i = \{(q_{t+1}, an_{t+1}),(q_{t+2}, an_{t+2}), ...,(q_{n-1}, an_{n-1})\}$ represents the ground-truth responses to achieve the specified outcome $T_i$.
The task of CCQG is formalized as follows: given a predefined outcome $T_i$, a context $C_i$, and a conversation history $H_i$, the objective of CCQG is to proactively guide the conversation to generate desired responses\footnote{In our paper, ``response'' refers to a question-answer pair. However, we sometimes use ``question'' interchangeably with ``response'' since the context frequently makes the answer evident once the question is posed.} $\mathcal{\hat{G}}_i$, thereby reaching the predefined outcome $T_i$.

\section{Methodology}
\label{sec:method}

\begin{figure*}[!t]
\centering 
\includegraphics[width=1\textwidth,height=2.0\textheight,keepaspectratio]{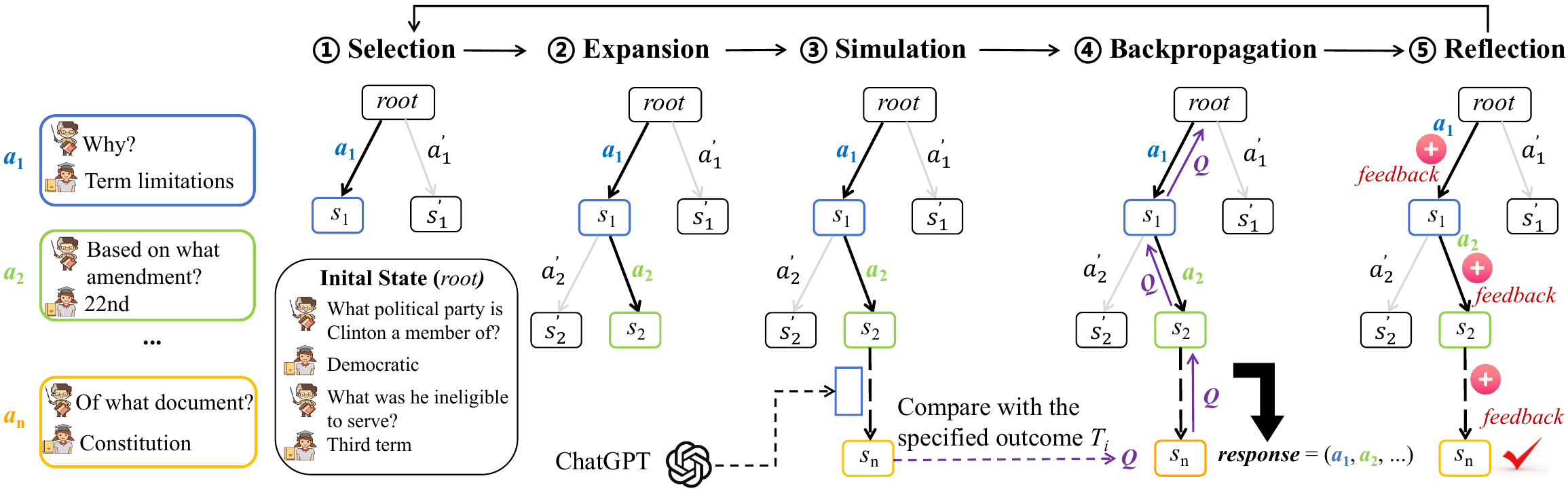}
\caption{An overview of our PCQPR framework. Specifically, the framework employs MCTS-like for planning, performing a lookahead search to generate the desired response (marked in \textbf{\textcolor{black}{black}}). Subsequently, it leverages a comparable reflection strategy to elevate the quality of this response (marked in \textbf{\textcolor{red}{red}}).
}
\label{fig:framework} 
\end{figure*}


\subsection{Overview}
To address the CCQG task, we propose a novel framework called PCQPR, designed to proactively steer the conversation towards achieving the specified outcome. 
The framework initially employs the Monte Carlo Tree Search-like (MCTS-like) algorithm for planning, performing a lookahead search to generate the desired response. Subsequently, it leverages the comparable reflection strategy to elevate the quality of this response.
The framework details are illustrated in Figure~\ref{fig:framework}. 
In planning, we innovatively integrate MCTS-like algorithms with LLMs, significantly narrowing the search space through efficient exploration.
Concurrently, we prioritize achieving the specified outcome as the primary objective, thereby enhancing the overall effectiveness of our planning methodology (see Section~\ref{sec: planning}).
In the comparable reflection strategy, we utilize LLMs to iteratively generate verbal feedback for each action along the planning path, facilitating learning from past failures and successful experiences. This continuous refinement enhances response quality, ensuring the achievement of the specified outcome (see Section~\ref{sec: reflection}).

\subsection{Planning with Monte Carlo Tree Search}
\label{sec: planning}
We explain how to (1) formulate the CCQG task as a Markov decision process \textbf{(MDP)} and (2) solve it by
\textbf{MCTS-like planning} algorithm.

\subsubsection{MDP}
For the CCQG task, we utilize the LLM (\emph{e.g.}, ChatGPT) to generate an action for the current state, thereby obtaining the next state.
Specifically, a state $s$ is the concatenation of the context $C_i$, the conversation history $H_i$, and the partially or completely generated response $\mathcal{\hat{G}}_i$, where the complete response signifies the attainment of the designated outcome $T_i$. 
An action $a$ comprises a question-answer pair that forms the basis of the complete response.
The terminal action matches or closely mirrors the predefined outcome, indicating the achievement of the desired goal.
The transition function systematically combines a state $s$ with an action $a$. An episode concludes once the LLM executes the terminal action.
To accurately evaluate the state $s$, the reward for $s$ is determined by how closely the responses reach a specified outcome, given that $s$ matches or closely resembles the predefined outcome~\footnote{We calculate the reward by measuring the semantic similarity between the predefined outcome and the terminal action using embeddings from SimCSE~\cite{SimCSE}. In this paper, we set the semantic similarity threshold at 0.6.}, \emph{i.e.}, REWARD(response) (line 17 in Algorithm~\ref{alg:mcts}). If the response fails to meet the specified outcome, the reward is set to 0.

To determine the optimal policy in an MDP, we explore a tree search-based planning algorithm, drawing inspiration from MCTS~\cite{GDP-Zero, mcts2, mcts3}.
More concretely, the tree search algorithm, inherently structured like a tree, assigns nodes to represent various states and uses edges to indicate actions. This algorithm begins at the root node, which is the initial state, and it explores the state space to identify terminal states with high rewards.
Each node in the tree search algorithm encompasses two essential elements: \textit{the visited number} and \textit{a value function}.
The visited number indicates how many times each node has been visited, while the value function represents the maximum reward obtained by starting from the node (or state)~\footnote{A node is defined to represent the state it embodies.} $s$ and executing action $a$.
The algorithm effectively maintains a proper balance between exploration and exploitation.
In the subsequent portion of this section, we detail the method of incorporating the tree search algorithm into the planning procedure.

\subsubsection{MCTS-like Planning}
To effectively guide the conversation toward a specified outcome, we develop an LLMs-based planning algorithm, which utilizes an MCTS-like tree search performing lookahead planning. 
We summarize the entire procedure in Algorithm~\ref{alg:mcts} and illustrate the process in Figure~\ref{fig:framework}.
More specifically, the planning process encompasses five critical operations: \textit{selection}, \textit{expansion}, \textit{simulation}, \textit{backpropagation}, and \textit{comparable reflection}.

\vpara{Selection.} 
This phase selects a branch of the tree for further exploration. Starting from the root node, representing the initial state $s$, the algorithm sequentially selects a child node until it reaches a leaf node.
To balance between exploration (\textit{focusing on less-visited nodes}) and exploitation (\textit{targeting high-value nodes}), we employ a function similar to the widely recognized Upper Confidence Bounds (UCB)~\cite{UCB} function to select each child node, which we refer to as UCB-like.
Unlike the UCB function, we use the probability of reaching the designated outcome for the exploitation term.
Formally, UCB-like function is defined as :
\begin{equation}
\label{eq:UCT}
    \text{UCB-like}(s) = Q(s, a) + w\sqrt{\frac{\log N(p)}{N(s)}}, 
\end{equation}
\noindent where $Q(s, a)$ is the maximum reward obtained by starting in $s$ and taking action $a$, $N(p)$ is the number of visits for parent node $p$ of $s$, $N(s)$ is the number of visits to a node $s$ , and $w$ is the exploration weight. 
At each level, the child node with the highest UCB-like value is selected.

\vpara{Expansion.}
From the node selected in the selection step, we derive $k$ possible actions and generate $k$ new states. These states form new nodes added to the list of children. More specifically, we use powerful LLMs (\emph{e.g.}, ChatGPT) to identify the most probable next actions, formalized as the function TOP\_K($s, k$), which returns the $k$ likeliest actions from state $s$. The corresponding $k$ new states are created by combining the current state with each action, and these new states are added to the list of children of the current node.

\vpara{Simulation.}
We calculate the expected reward (\emph{i.e.}, $Q$ value) by leveraging LLMs (\emph{e.g.}, ChatGPT) to simulate potential scenarios from the current node. Starting at the current node, LLMs generate actions sequentially until a terminal action is reached or a maximum number of steps is completed. We then assess the semantic similarity between the final action and the specified outcome $T_i$. This similarity serves as the reward for the current node, measuring its relevance to the outcome.

\vpara{Backpropagation.}
Following the simulation phase, we can obtain a path sequence of actions from the root node to the terminal node, denoted as $response$, and acquire the corresponding reward value associated with this path.
Subsequently, this reward is backpropagated to revise the $Q$ value of each state-action pair encountered along the path. 
Ultimately, the $Q$  value of every node in this path is updated to reflect the simulation results accurately.

\begin{algorithm}[!t]	
\small
\renewcommand{\algorithmicrequire}{\textbf{Input:}}
\renewcommand{\algorithmicensure}{\textbf{Output:}}
	\renewcommand{\algorithmicreturn}{\textbf{Return}}
	\caption{\textbf{The PCQPR algorithm}}
	\label{alg:mcts}
	\begin{algorithmic}[1]
	    \REQUIRE The initial state $s$ (or $root$), the number of children for each node $k$, UCB-like exploration parameter $w$.
	    \ENSURE Response with the highest reward.
	    \STATE Initialize the $response\_dict$ = OrderedDict()
	    \FOR{$i \leftarrow$ 1, 2, ..., $max\_iterations$}
	    \STATE $node \leftarrow root$
	    \STATE \textcolor{blue}{\# Selection}
            \WHILE{$ \left |node.children \right|$ > 0}
            \STATE $node \leftarrow \text{UCB-like}(node.children)$
            \ENDWHILE
            \STATE \textcolor{blue}{\# Expansion}
            \STATE $next\_actions\leftarrow \text{TOP\_K}(node, k)$ 
            \FOR{$next\_action \in next\_actions$}
            \STATE $next\_state \leftarrow \text{COMBINE}(node, next\_action)$
           \STATE $new\_node \leftarrow new\_state$
            \STATE Append $new\_node$ to children of $node$
            \ENDFOR
            \STATE \textcolor{blue}{\# Simulation}
            \STATE $response \leftarrow \text{LLM}(node)$
            \STATE $reward \leftarrow \text{REWARD}(response)$
            \STATE $response\_dict\text{[}response\text{]} = reward$
            \STATE \textcolor{blue}{\# Backpropagation}
            \STATE Update the values of $node$ and its ancestors with $reward$
             \STATE \textcolor{blue}{\# Reflection}
            \STATE $feedback \leftarrow \text{COMP\_REFINE}(response)$
            \STATE Add the $feedback$ into $node$ and its ancestors
	    \ENDFOR      
     \RETURN $response$ with the highest reward in $response\_dict$.
	\end{algorithmic} 

\end{algorithm}

\subsection{Comparable Reflection}
\label{sec: reflection}
Although the MCTS-like planning algorithm demonstrates commendable performance in the CCQG task, there is significant potential for further improvement. Notably, some initial planning paths fail to achieve the desired outcomes, underscoring the need for a more refined approach to enhance the algorithm's effectiveness. Inspired by human cognitive strategies, which learn from past \textit{successes} and \textit{failures} to improve future performance,  we propose an enhancement for the MCTS-like planning algorithm through a comparable reflection mechanism that summarizes both errors and successful experiences. This mechanism provides the algorithm with detailed verbal feedback on previous planning paths, enabling it to optimize future decision-making processes.

To implement this enhancement, we integrate a self-reflection mechanism powered by advanced LLMs (\emph{e.g.}, ChatGPT). This marks a significant advancement in the MCTS-like planning algorithm's effectiveness for the CCQG task. The sophisticated analytical capabilities of LLMs~\cite{Liang_ACL24} enable comprehensive evaluations of planning paths, generating detailed verbal reflections for each sequence of actions (\emph{i.e.}, COMP\_REFINE(response), see line 22 in Alogrithm~\ref{alg:mcts}). 
These reflections serve as a semantic guidance signal, providing the algorithm with concrete directions for improvement by pinpointing failure points and highlighting successful strategies. This balanced view of what works and what doesn’t fosters a deeper comprehension of the decision-making dynamics, enabling the detection of patterns that lead to both suboptimal results and successful outcomes. The algorithm can be finely tuned by systematically analyzing these patterns to anticipate and circumvent similar pitfalls while replicating successful strategies in future iterations. This reflective process provides valuable insights that optimize the algorithm’s behavior, enhancing its ability to solve complex scenarios through iterative trials and self-reflection. 

For example, in planning, there are two initial planning paths: a failed path $p_1 = (a_1, ..., a_n)$ where the terminal action $a_n$ diverges from the outcome $T_i$, and a successful path $p_2 = (a_1^{\prime}, ..., a_n^{\prime})$ where the terminal action $a_n^{\prime}$ matches the outcome $T_i$. 
The comparable reflection mechanism provides fine-grained feedback for two planning paths.
For the failed path $p_1$, it identifies the points of divergence and reasons for failure, providing critical feedback on what went wrong. Conversely, for the successful path $p_2$, it recognizes the key decisions that led to a positive outcome, reinforcing effective strategies. This dual analysis ensures that the algorithm not only learns from its mistakes but also builds on its successes, leading to continuous improvement and more reliable planning outcomes. This comparable reflection mechanism ensures that future iterations of the planning algorithm are better equipped to handle similar situations, ultimately enhancing performance and decision-making capabilities in complex scenarios. 
To the best of our knowledge, we are the first to combine a tree search algorithm with a comparable reflection strategy to enhance the performance of the novel CCQG task.

\section{Experiments}
\label{sec:experiment}

\subsection{Experimental Settings}
\label{sec:exp_settings}

\begin{table*}[!t]
\small
\centering
\newcolumntype{?}{!{\vrule width 1pt}}
\renewcommand\arraystretch{1}
\scalebox{1.0}{
\begin{tabular}{@{}c@{ }?ccccc@{}}
\toprule
\textbf{Model} & \textbf{BLEU}& \textbf{METEOR}& \textbf{Conv-last1} & \textbf{Conv-last2}& \textbf{Success Rate}\\
\midrule
SG-CQG & 3.62 & 46.35  & 32.77 & 40.04  & 15.40 \\
COT & 3.86 &46.62 & 31.39 & 39.48 & 19.20 \\
COT-SC & 6.85& 53.45& 30.84 & 38.86 & 10.40 \\
TOT & \textbf{7.05} & \textbf{54.81} & 29.85 & 37.96 & 23.20\\
Mixtral-8x7B & 1.96 & 46.93  & 29.26 & 36.65 & 6.20\\
ChatGPT & 3.31 & 45.57 & 31.09 & 39.19 & 19.80\\
GPT-4-Turbo & 5.28  & 52.48 & 30.29 & 37.53 & 12.80\\
\midrule
PCQPR(Mixtral-8x7B)& 1.54 & 44.87 & 34.78  &40.62  & 25.80\\
PCQPR(ChatGPT)& 3.56  & 50.07 & 36.13 & 42.31 & 28.00\\
PCQPR(GPT-4-Turbo)& 6.47 & 54.04  & \textbf{37.39} & \textbf{43.06} & \textbf{35.00}\\
\bottomrule
\end{tabular}
}
\caption{Overall evaluation on CoQA (\%).}
\label{tab:overall_evaluation}
\end{table*}

\begin{table*}[!t]
\small
\centering
\newcolumntype{?}{!{\vrule width 1pt}}
\renewcommand\arraystretch{1}
\scalebox{1}{
\begin{tabular}{@{}c@{ }?ccccc@{}}
\toprule
\textbf{Model} & \textbf{BLEU}& \textbf{METEOR}& \textbf{Conv-last1} & \textbf{Conv-last2}& \textbf{Success Rate}\\
\midrule
COT & 11.17 &41.87 & 46.13 &50.86  & 30.32 \\
COT-SC &\textbf{23.38} &\textbf{51.82} & 46.79 &51.09  & 22.62 \\
TOT & 13.74 &46.09 & 48.19 &52.48  &40.27 \\
Mixtral-8x7B & 7.45 & 39.77  &44.75  &49.68 &23.98 \\
ChatGPT &11.02  &41.29  & 46.04 &50.33  & 27.60\\
GPT-4-Turbo &7.85  &40.46 &44.55  &49.42  &26.24 \\
\midrule
PCQPR(Mixtral-8x7B)&5.64  &38.84 & 51.85 &56.11  & 63.35\\
PCQPR(ChatGPT)&9.38  & 41.41 &52.87 &56.06  &67.42 \\
PCQPR(GPT-4-Turbo)&7.59 & 42.47 & \textbf{54.54} &\textbf{57.49}  & \textbf{70.59}\\
\bottomrule
\end{tabular}
}
\caption{Overall evaluation on QuAC (\%).}
\label{tab:overall_evaluation2}
\end{table*}

\vpara{Datasets.} 
As the novel task setting of CCQG has not been investigated in prior research, no direct dataset is available. Hence, we modify two popular conversational datasets (\emph{i.e.}, CoQA~\cite{CoQA} and QuAC~\cite{QuAC}) to meet the requirements of the CCQG task.
For the CoQA dataset, we standardize the number of turns in all conversations to 10 because some have too many turns. For the QuAC dataset, we include only conversations where the turn is labeled with `followup: y' or `followup: m', resulting in a training set of 2,749 conversations and a validation set of 221 conversations. We use the last three question-answer pairs as the target pairs.
The two datasets comprise conversations across a wide range of domains, where each conversation consists of a relevant context and several question-answer pairs.
Following prior studies~\cite{SG-CQG}, we use the validation sets from two datasets as our test sets since the original test sets are unavailable.

\hide{
We evaluate our proposed approach on CoQA~\cite{CoQA}, a large-scale CQA dataset created on Amazon Mechanical Turk. CoQA is the most widely used dataset for the conversation question generation task.
CoQA contains about 7,699 conversations from seven diverse domains. Each conversation is composed of a related context and multiple question-answer pairs, resulting in a total of 127,000 question-answer pairs.
More specifically, CoQA encompasses a training set with 7,199 conversations and a validation set with 500 conversations.
Following previous work~\cite{SG-CQG}, we use the validation set of CoQA as our test set because its test set is unavailable.}

\vpara{Automatic Evaluation Metrics.} 
We employ widely used evaluation metrics for text generation, including BLEU~\cite{bleu} and METEOR~\cite{meteor}, to evaluate the generated response. 
Concretely, BLEU evaluates the precision of the generated text compared to the reference text.
METEOR is a detailed evaluation considering exact words, synonyms, and similar phrases.
Additionally, we employ SimCSE~\cite{SimCSE} to measure conversational coherence. This method regards the generated response as the premise, while the previous conversational history serves as the hypothesis. It then calculates a similarity score between them to assess topic coherence. In our experiments, we assess the conversation coherence score by focusing on the last one and last two prior turns, referred to as Conv-last1 and Conv-last2, respectively.
Furthermore, the most pivotal metric is the Success Rate, \emph{i.e.}, Success Rate = $\frac{\text{Number of Successful Outcomes}}{\text{Total Number of Instances}}$, which is employed to measure the capacity to achieve the intended conversational outcome.

\vpara{Human Evaluation Metrics.} 
Due to the extensive critique of automatic metrics for their poor alignment with human assessments~\cite{nlgmetrics}, we incorporate two human evaluation metrics: Coherence and Effectiveness. 
The former metric evaluates whether the entire conversation maintains logical and topic coherence, while the latter assesses the efficiency with which the intended conversational outcome is achieved.
We randomly select 50 conversations from the validation set of the CoQA dataset and invite three individuals to score the generated responses, with a scoring range of \{0, 1, 2\}, where a higher score indicates better quality in terms of Coherence and Effectiveness. Detailed scoring criteria are provided in Appendix~\ref{sec:append_human}.
We use Fleiss's kappa~\cite{fleiss1971measuring} to measure the consistency among the three persons.

\vpara{Baselines.} 
Since this novel task setting has not been fully explored in previous work, there are no readily available baselines for comparison. Therefore, we have modified the seven most relevant baselines for comparison.
Among them, SG-CQG~\cite{SG-CQG}, an advanced CQG model assessed on the CoQA dataset, proposes a two-stage method including a `what-to-ask' module and a `how-to-ask' module. 
The most popular open-source model, Mixtral-8x7B~\cite{mixtral}, along with two closed-source models, ChatGPT\footnote{https://openai.com/blog/chatgpt} and GPT-4-Turbo~\cite{GPT-4}, are employed in a zero-shot setting to address this task.
COT~\cite{COT} and its variants, COT-SC~\cite{COT-SC} and TOT~\cite{TOT} utilize ChatGPT as the backbone to solve this task.

\subsection{Overall Evaluation}
Table~\ref{tab:overall_evaluation} and Table~\ref{tab:overall_evaluation2} show the overall evaluation results for the CoQA and QuAC datasets, we can conclude that: \textbf{(1) Our proposed framework, PCQPR, can produce more logically coherent responses.} Compared with baselines, our approaches (\emph{e.g.}, PCQPR(ChatGPT) and PCQPR(GPT-4-Turbo)) excel in generating more coherent responses that align with the conversational history (Conv-last1 and Conv-last2), indicating their impressive abilities in conversational understanding and response generation.
For instance, on the CoQA dataset, PCQPR(GPT-4-Turbo) achieves absolute improvements of 4.62\% in Conv-last1 and 3.02\% in Conv-last2 over the best baseline (\emph{i.e.}, SG-CQG).
\textbf{(2) Our approach naturally transitions into the specified outcome with the highest success rate, demonstrating its effectiveness.} 
We observe that our best method (\emph{i.e.}, PCQPR(GPT-4-Turbo)) achieves an 11.8\% improvement in Success Rate over the best baseline (\emph{i.e.}, TOT) on CoQA.
Additionally, on the CoQA dataset, PCQPR(GPT-4-Turbo) derives a 22.2\% Success Rate gain over its corresponding vanilla model GPT-4-Turbo.
PCQPR(ChatGPT) obtains an 8.2\% Success Rate gain over its corresponding vanilla model ChatGPT.
\textbf{(3) Our approach not only produces logically coherent responses but also effectively steers the conversation toward the predetermined outcome.} PCQPR(GPT-4-Turbo) derives 37.39\% Conv-last1, 43.06\% Conv-last2, and 35.00\% Success Rate on CoQA.
Compared to the best baseline results, SG-CQG exhibits the highest coherence, achieving 32.77\%  Conv-last1 and 40.04\% Conv-last2, while TOT shows the highest Success Rate at 23.20\%. Our proposed framework demonstrates superiority in both aspects.
\textbf{(4) Our method has comparable performance in BLEU and METEOR with the best baseline.} We observe that the best baseline TOT derives 7.05\% BLEU and 54.81\%, while our method PCQPR(GPT-4-Turbo) achieves 6.47\% BLEU and 54.04\% on CoQA. 
Although there are gaps, this does not mean that our approach is inferior to the baseline.
Because BLEU and METEOR metrics are less reliable, they measure the lexical surface similarity between the produced response and the ground-truth response.
But conversations on the same topic can be described in different ways, rather than just ground-truth responses.
Therefore, we focus primarily on reliable metrics of conversational coherence (\emph{i.e.}, Conv-last1 and Conv-last2) and effectiveness (\emph{i.e.}, Success Rate).

\subsection{Human Evaluation}
To comprehensively validate the effectiveness of our approach, we conduct a rigorous human evaluation. This involves providing scorers with detailed examples, designed to guide them towards objective and fair assessments of generated responses. The results presented in Table~\ref{tab:human evaluation} demonstrate the superiority of our method compared to traditional baselines.
Notably, our approach shows superiority in terms of conversational coherence (\emph{i.e.}, ``Coherence'') and in successfully reaching the designated conversational outcome (\emph{i.e.}, ``Effectiveness''). These two aspects are crucial in assessing the quality of conversational question-answering systems, as they directly reflect the system's ability to maintain a natural and purposeful conversation flow. 
To ensure the reliability of these evaluation results, we further employ Fleiss's kappa, a statistical measure designed to assess the consistency of agreement among multiple scorers. 
The kappa values range between 0.41 and 0.60, indicating moderate agreement among the three scorers. This range suggests a reasonable consensus,  lending further credibility to our results.


\begin{table}[!t]
\centering
\small
\newcolumntype{?}{!{\vrule width 1pt}}
\renewcommand\arraystretch{1.0}
\scalebox{1.0}{
\begin{tabular}{@{}c@{ }?@{ }cc@{ }}
\toprule
\textbf{Model} & \textbf{Coherence} & \textbf{Effectiveness} \\
\midrule
SG-CQG &1.28 &1.14 \\
TOT& 1.16 & 1.25\\
Mixtral-8x7B  & 1.12 & 0.91\\
ChatGPT& 1.23&1.20  \\
GPT-4-Turbo  & 1.18& 1.06 \\
\midrule
PCQPR(Mixtral-8x7B) & 1.33 &1.29 \\
PCQPR(ChatGPT) & 1.65 &1.34 \\
PCQPR(GPT-4-Turbo) & 1.68 & 1.62\\
\midrule
kappa & 0.46&0.53 \\
\bottomrule
\end{tabular}
}
\caption{Human evaluation results on CoQA.
}
\label{tab:human evaluation}
\end{table}

\begin{table}[!t]
\centering
\small
\newcolumntype{?}{!{\vrule width 1pt}}
\renewcommand\arraystretch{1.0}
\scalebox{1.0}{
\begin{tabular}{@{}c@{ }?@{ }cc@{ }}
\toprule
\textbf{Model} & \textbf{Conv-last1} & \textbf{Success Rate} \\
\midrule
PCQPR(Mixtral-8x7B)  & 34.78&25.80 \\
- w/o MCTS & 28.69 & 14.60\\
- w/o Reflection & 34.80  & 15.40 \\
\midrule
PCQPR(ChatGPT) & 36.13 & 28.00\\
- w/o MCTS & 30.16 & 26.80\\
- w/o Reflection & 36.18 & 22.00 \\
\midrule
PCQPR(GPT-4-Turbo) & 37.39 & 35.00\\
- w/o MCTS & 29.03 & 14.60\\
- w/o Reflection & 36.85 & 30.60\\
\bottomrule
\end{tabular}
}
\caption{Ablation studies for PCQPR on CoQA (\%).}
\label{tab:ablation_study}
\end{table}

\subsection{Ablation Studies}
To verify the effectiveness of PCQPR, we conduct extensive ablation experiments.

\subsubsection{Effect of MCTS-based Planner}
To investigate the effectiveness of our proposed MCTS-based planner, we conduct an experiment where the MCTS-based planner is removed, denoted as ``w/o MCTS''.
Table~\ref{tab:ablation_study} reports the results on Conv-last1 and Success Rate.
We observe that removing the MCTS-based planner results in a reduction of 8.36\% in Conv-last1 score and 20.4\% in Success Rate for our method, PCQPR(GPT-4-Turbo).
A reasonable explanation is that the MCTS-based planner performs a lookahead search and finds high-quality responses toward the predefined outcome.
Consequently, the MCTS-based planner we designed plays a crucial role in our framework.

\subsubsection{Effect of Reflection Strategy}

We evaluate the impact of the reflection strategy on our proposed approach. This involves a contrastive analysis, termed ``w/o Reflection'', specifically designed to measure the contribution of this strategy. 
As shown in Table~\ref{tab:ablation_study}, the results indicate a significant performance decrease in the absence of reflection: a 10.4\% drop in Success Rate for PCQPR(Mixtral-8x7B). 
This decline underscores the reflection strategy's vital role, in which verbal feedback integration facilitates an iterative learning process.
This process enables the model to refine its response generation based on previous outcomes, improving relevance and accuracy. 
Therefore, the reflection strategy elevates performance and marks a significant advancement in conversational question-answering systems. It enhances the model's adaptability and effectiveness, particularly in dynamic and evolving conversational contexts, thereby underscoring its importance in developing more sophisticated, responsive systems.

\section{Conclusion}
\label{sec:conclusion}
We present a novel task setting, CCQG, designed to generate subsequent questions that proactively guide conversations toward the specified outcome.
To address this task, we propose an innovative framework, PCQPR, which uniquely combines the MCTS-like planning algorithm with LLMs to enhance planning capabilities. This approach conducts a lookahead search to explore multiple potential paths.
Furthermore, we introduce a novel reflection mechanism that provides insightful verbal feedback for each action along the entire planning path.
Extensive experiments demonstrate the superior performance of PCQPR over closed-source and open-source LLMs.
We believe this effort could be inspiring for future research in AI-driven conversational question-answering systems.
\section*{Acknowledgments}
This work is supported by the National Key Research \& Develop Plan (2023YFF0725100) and the National Natural Science Foundation of China (62322214, U23A20299, 62076245, 62072460, 62172424, 62276270). This work is supported by Public Computing Cloud, Renmin University of China.  We also acknowledge the support from the China Scholarship Council Scholarship Fund. We sincerely appreciate the valuable and insightful feedback provided by all reviewers.
\section*{Limitations}
 Despite the effectiveness of the comparable self-reflection strategy, it still exhibits certain limitations. This paper primarily focuses on leveraging large language models (LLMs) to provide valuable verbal feedback and refine each action within the entire planning path. However, we recognize that the inherent capabilities of LLMs may constrain their effectiveness. In particular, for some paths that initially succeed in reaching the specific outcome, the reflection strategy may not yield further improvements. In future work, we will explore optimal methods for integrating human feedback with verbal reflection generated by LLMs to address this challenge above effectively.
\bibliography{anthology}

\begin{thebibliography}{37}
\expandafter\ifx\csname natexlab\endcsname\relax\def\natexlab#1{#1}\fi

\bibitem[{Achiam et~al.(2023)Achiam, Adler, Agarwal, Ahmad, Akkaya, Aleman, Almeida, Altenschmidt, Altman, Anadkat et~al.}]{GPT-4}
Josh Achiam, Steven Adler, Sandhini Agarwal, Lama Ahmad, Ilge Akkaya, Florencia~Leoni Aleman, Diogo Almeida, Janko Altenschmidt, Sam Altman, Shyamal Anadkat, et~al. 2023.
\newblock Gpt-4 technical report.
\newblock \emph{arXiv preprint arXiv:2303.08774}, pages 1--100.

\bibitem[{Banerjee and Lavie(2005)}]{meteor}
Satanjeev Banerjee and Alon Lavie. 2005.
\newblock {METEOR:} an automatic metric for {MT} evaluation with improved correlation with human judgments.
\newblock In \emph{Proceedings of the Workshop on Intrinsic and Extrinsic Evaluation Measures for Machine Translation and/or Summarization@ACL 2005}, pages 65--72.

\bibitem[{Chen et~al.(2023)Chen, Wu, and Zaki}]{chen2020toward}
Yu~Chen, Lingfei Wu, and Mohammed~J. Zaki. 2023.
\newblock Toward subgraph-guided knowledge graph question generation with graph neural networks.
\newblock \emph{{IEEE} Transactions on Neural Networks and Learning Systems}, pages 1--12.

\bibitem[{Choi et~al.(2018)Choi, He, Iyyer, Yatskar, Yih, Choi, Liang, and Zettlemoyer}]{QuAC}
Eunsol Choi, He~He, Mohit Iyyer, Mark Yatskar, Wen{-}tau Yih, Yejin Choi, Percy Liang, and Luke Zettlemoyer. 2018.
\newblock Quac: Question answering in context.
\newblock In \emph{Proceedings of the 2018 Conference on Empirical Methods in Natural Language Processing, {EMNLP} 2018}, pages 2174--2184.

\bibitem[{Dao et~al.(2023)Dao, Liao, Le, and Nie}]{dao2023reinforced}
Huy Dao, Lizi Liao, Dung Le, and Yuxiang Nie. 2023.
\newblock Reinforced target-driven conversational promotion.
\newblock In \emph{Proceedings of the 2023 Conference on Empirical Methods in Natural Language Processing}, pages 12583--12596.

\bibitem[{Deng et~al.(2022)Deng, Lei, Zhang, Lam, and Chua}]{AskClarify}
Yang Deng, Wenqiang Lei, Wenxuan Zhang, Wai Lam, and Tat{-}Seng Chua. 2022.
\newblock {PACIFIC:} towards proactive conversational question answering over tabular and textual data in finance.
\newblock In \emph{Proceedings of the 2022 Conference on Empirical Methods in Natural Language Processing, {EMNLP} 2022}, pages 6970--6984.

\bibitem[{Do et~al.(2023)Do, Zou, Joty, Tai, Pan, Chen, and Aw}]{SG-CQG}
Xuan~Long Do, Bowei Zou, Shafiq~R. Joty, Anh~Tran Tai, Liangming Pan, Nancy~F. Chen, and Ai~Ti Aw. 2023.
\newblock Modeling what-to-ask and how-to-ask for answer-unaware conversational question generation.
\newblock In \emph{Proceedings of the 61st Annual Meeting of the Association for Computational Linguistics, {ACL} 2023}, pages 10785--10803.

\bibitem[{Do et~al.(2022)Do, Zou, Pan, Chen, Joty, and Aw}]{CQG}
Xuan~Long Do, Bowei Zou, Liangming Pan, Nancy~F. Chen, Shafiq~R. Joty, and Ai~Ti Aw. 2022.
\newblock Cohs-cqg: Context and history selection for conversational question generation.
\newblock In \emph{Proceedings of the 29th International Conference on Computational Linguistics, {COLING} 2022}, pages 580--591.

\bibitem[{Fei et~al.(2022)Fei, Zhang, Gui, Liang, Wang, Wu, and Huang}]{ControllQG}
Zichu Fei, Qi~Zhang, Tao Gui, Di~Liang, Sirui Wang, Wei Wu, and Xuanjing Huang. 2022.
\newblock {CQG:} {A} simple and effective controlled generation framework for multi-hop question generation.
\newblock In \emph{Proceedings of the 60th Annual Meeting of the Association for Computational Linguistics, {ACL} 2022}, pages 6896--6906.

\bibitem[{Fleiss(1971)}]{fleiss1971measuring}
Joseph~L Fleiss. 1971.
\newblock Measuring nominal scale agreement among many raters.
\newblock \emph{Psychological bulletin}, 76(5):378--382.

\bibitem[{Gao et~al.(2021)Gao, Yao, and Chen}]{SimCSE}
Tianyu Gao, Xingcheng Yao, and Danqi Chen. 2021.
\newblock Simcse: Simple contrastive learning of sentence embeddings.
\newblock In \emph{Proceedings of the 2021 Conference on Empirical Methods in Natural Language Processing, {EMNLP} 2021}, pages 6894--6910.

\bibitem[{Gou et~al.(2023)Gou, Xia, Yu, Yu, Huang, Li, and Nguyen}]{DiversifyQG}
Qi~Gou, Zehua Xia, Bowen Yu, Haiyang Yu, Fei Huang, Yongbin Li, and Cam{-}Tu Nguyen. 2023.
\newblock Diversify question generation with retrieval-augmented style transfer.
\newblock In \emph{Proceedings of the 2023 Conference on Empirical Methods in Natural Language Processing, {EMNLP} 2023}, pages 1677--1690.

\bibitem[{Gu et~al.(2021)Gu, Mirshekari, Yu, and Sisto}]{ChainCQG}
Jing Gu, Mostafa Mirshekari, Zhou Yu, and Aaron Sisto. 2021.
\newblock Chaincqg: Flow-aware conversational question generation.
\newblock In \emph{Proceedings of the 16th Conference of the European Chapter of the Association for Computational Linguistics, {EACL} 2021}, pages 2061--2070.

\bibitem[{Guo et~al.(2024{\natexlab{a}})Guo, Liao, Li, and Chua}]{NQG_Survey}
Shasha Guo, Lizi Liao, Cuiping Li, and Tat{-}Seng Chua. 2024{\natexlab{a}}.
\newblock A survey on neural question generation: Methods, applications, and prospects.
\newblock \emph{CoRR}, abs/2402.18267.

\bibitem[{Guo et~al.(2024{\natexlab{b}})Guo, Liao, Zhang, Wang, Li, and Chen}]{SGSH}
Shasha Guo, Lizi Liao, Jing Zhang, Yanling Wang, Cuiping Li, and Hong Chen. 2024{\natexlab{b}}.
\newblock {SGSH:} stimulate large language models with skeleton heuristics for knowledge base question generation.
\newblock In \emph{Findings of the Association for Computational Linguistics: {NAACL} 2024}, pages 4613--4625.

\bibitem[{Guo et~al.(2024{\natexlab{c}})Guo, Zhang, Ke, Li, and Chen}]{Diversify_Guo}
Shasha Guo, Jing Zhang, Xirui Ke, Cuiping Li, and Hong Chen. 2024{\natexlab{c}}.
\newblock Diversifying question generation over knowledge base via external natural questions.
\newblock In \emph{Proceedings of the 2024 Joint International Conference on Computational Linguistics, Language Resources and Evaluation, {LREC/COLING} 2024}, pages 5096--5108.

\bibitem[{Guo et~al.(2022)Guo, Zhang, Wang, Zhang, Li, and Chen}]{DSM}
Shasha Guo, Jing Zhang, Yanling Wang, Qianyi Zhang, Cuiping Li, and Hong Chen. 2022.
\newblock Dsm: Question generation over knowledge base via modeling diverse subgraphs with meta-learner.
\newblock In \emph{Proceedings of the 2022 Conference on Empirical Methods in Natural Language Processing, {EMNLP} 2022}, pages 4194--4207.

\bibitem[{Hao et~al.(2023)Hao, Gu, Ma, Hong, Wang, Wang, and Hu}]{RAP}
Shibo Hao, Yi~Gu, Haodi Ma, Joshua~Jiahua Hong, Zhen Wang, Daisy~Zhe Wang, and Zhiting Hu. 2023.
\newblock Reasoning with language model is planning with world model.
\newblock In \emph{Proceedings of the 2023 Conference on Empirical Methods in Natural Language Processing, {EMNLP} 2023}, pages 8154--8173.

\bibitem[{He et~al.(2024)He, Liao, Cao, Liu, Liu, Chen, and Qin}]{he-etal-2024-planning}
Tao He, Lizi Liao, Yixin Cao, Yuanxing Liu, Ming Liu, Zerui Chen, and Bing Qin. 2024.
\newblock Planning like human: A dual-process framework for dialogue planning.
\newblock In \emph{Proceedings of the 62nd Annual Meeting of the Association for Computational Linguistics}, pages 4768--4791.

\bibitem[{Jiang et~al.(2024)Jiang, Sablayrolles, Roux, Mensch, Savary, Bamford, Chaplot, Casas, Hanna, Bressand et~al.}]{mixtral}
Albert~Q Jiang, Alexandre Sablayrolles, Antoine Roux, Arthur Mensch, Blanche Savary, Chris Bamford, Devendra~Singh Chaplot, Diego de~las Casas, Emma~Bou Hanna, Florian Bressand, et~al. 2024.
\newblock Mixtral of experts.
\newblock \emph{arXiv preprint arXiv:2401.04088}, pages 1--13.

\bibitem[{Kocsis and Szepesv{\'{a}}ri(2006)}]{UCB}
Levente Kocsis and Csaba Szepesv{\'{a}}ri. 2006.
\newblock Bandit based monte-carlo planning.
\newblock In \emph{Proceedings of the 17th European Conference on Machine Learning, {ECML} 2006}, volume 4212, pages 282--293.

\bibitem[{Liang et~al.(2024)Liang, Liao, Fei, and Jiang}]{Liang_ACL24}
Jinggui Liang, Lizi Liao, Hao Fei, and Jing Jiang. 2024.
\newblock Synergizing large language models and pre-trained smaller models for conversational intent discovery.
\newblock In \emph{Findings of the Association for Computational Linguistics, {ACL} 2024}, pages 14133--14147.

\bibitem[{Liang et~al.(2023)Liang, Wang, Zhu, Wang, Qian, and Lan}]{PCQG}
Yuanyuan Liang, Jianing Wang, Hanlun Zhu, Lei Wang, Weining Qian, and Yunshi Lan. 2023.
\newblock Prompting large language models with chain-of-thought for few-shot knowledge base question generation.
\newblock In \emph{Proceedings of the 2023 Conference on Empirical Methods in Natural Language Processing, {EMNLP} 2023}, pages 4329--4343.

\bibitem[{Liao et~al.(2023)Liao, Yang, and Shah}]{liao2023proactive}
Lizi Liao, Grace~Hui Yang, and Chirag Shah. 2023.
\newblock Proactive conversational agents in the post-chatgpt world.
\newblock In \emph{Proceedings of the 46th International ACM SIGIR Conference on Research and Development in Information Retrieval, {SIGIR} 2023}, pages 3452--3455.

\bibitem[{Ling et~al.(2023)Ling, Cai, Liu, Chen, and de~Rijke}]{CQGTOIS}
Yanxiang Ling, Fei Cai, Jun Liu, Honghui Chen, and Maarten de~Rijke. 2023.
\newblock Generating relevant and informative questions for open-domain conversations.
\newblock \emph{{ACM} Trans. Inf. Syst.}, 41(1):1--30.

\bibitem[{Madaan et~al.(2023)Madaan, Tandon, Gupta, Hallinan, Gao, Wiegreffe, Alon, Dziri, Prabhumoye, Yang, Welleck, Majumder, Gupta, Yazdanbakhsh, and Clark}]{Self-Refine}
Aman Madaan, Niket Tandon, Prakhar Gupta, Skyler Hallinan, Luyu Gao, Sarah Wiegreffe, Uri Alon, Nouha Dziri, Shrimai Prabhumoye, Yiming Yang, Sean Welleck, Bodhisattwa~Prasad Majumder, Shashank Gupta, Amir Yazdanbakhsh, and Peter Clark. 2023.
\newblock Self-refine: Iterative refinement with self-feedback.
\newblock In \emph{Proceedings of 37th Conference on Neural Information Processing Systems, {NeurIPS} 2023}, pages 1--54.

\bibitem[{Novikova et~al.(2017)Novikova, Dusek, Curry, and Rieser}]{nlgmetrics}
Jekaterina Novikova, Ondrej Dusek, Amanda~Cercas Curry, and Verena Rieser. 2017.
\newblock Why we need new evaluation metrics for {NLG}.
\newblock In \emph{Proceedings of the 2017 Conference on Empirical Methods in Natural Language Processing, {EMNLP} 2017}, pages 2241--2252.

\bibitem[{Papineni et~al.(2002)Papineni, Roukos, Ward, and Zhu}]{bleu}
Kishore Papineni, Salim Roukos, Todd Ward, and Wei-Jing Zhu. 2002.
\newblock Bleu: a method for automatic evaluation of machine translation.
\newblock In \emph{Proceedings of the 40th annual meeting of the Association for Computational Linguistics, {ACL} 2002}, pages 311--318.

\bibitem[{Reddy et~al.(2019)Reddy, Chen, and Manning}]{CoQA}
Siva Reddy, Danqi Chen, and Christopher~D. Manning. 2019.
\newblock Coqa: {A} conversational question answering challenge.
\newblock \emph{Trans. Assoc. Comput. Linguistics}, 7:249--266.

\bibitem[{Shinn et~al.(2023)Shinn, Cassano, Gopinath, Narasimhan, and Yao}]{shinn2023reflexion}
Noah Shinn, Federico Cassano, Ashwin Gopinath, Karthik~R Narasimhan, and Shunyu Yao. 2023.
\newblock Reflexion: Language agents with verbal reinforcement learning.
\newblock In \emph{Proceedings of the 37th Conference on Neural Information Processing Systems, {NeurIPS} 2023}, pages 1--19.

\bibitem[{Wang et~al.(2023)Wang, Wei, Schuurmans, Le, Chi, Narang, Chowdhery, and Zhou}]{COT-SC}
Xuezhi Wang, Jason Wei, Dale Schuurmans, Quoc~V. Le, Ed~H. Chi, Sharan Narang, Aakanksha Chowdhery, and Denny Zhou. 2023.
\newblock Self-consistency improves chain of thought reasoning in language models.
\newblock In \emph{The Eleventh International Conference on Learning Representations, {ICLR} 2023}, pages 1--24.

\bibitem[{Wei et~al.(2022)Wei, Wang, Schuurmans, Bosma, Ichter, Xia, Chi, Le, and Zhou}]{COT}
Jason Wei, Xuezhi Wang, Dale Schuurmans, Maarten Bosma, Brian Ichter, Fei Xia, Ed~H. Chi, Quoc~V. Le, and Denny Zhou. 2022.
\newblock Chain-of-thought prompting elicits reasoning in large language models.
\newblock In \emph{Proceedings of the 36th Conference on Neural Information Processing Systems, {NeurIPS} 2022}, pages 1--14.

\bibitem[{Yao et~al.(2023)Yao, Yu, Zhao, Shafran, Griffiths, Cao, and Narasimhan}]{TOT}
Shunyu Yao, Dian Yu, Jeffrey Zhao, Izhak Shafran, Thomas~L. Griffiths, Yuan Cao, and Karthik Narasimhan. 2023.
\newblock Tree of thoughts: Deliberate problem solving with large language models.
\newblock In \emph{Proceedings of the 37th Conference on Neural Information Processing Systems, {NeurIPS} 2023}, pages 1--14.

\bibitem[{Yu et~al.(2023)Yu, Chen, and Yu}]{GDP-Zero}
Xiao Yu, Maximillian Chen, and Zhou Yu. 2023.
\newblock Prompt-based monte-carlo tree search for goal-oriented dialogue policy planning.
\newblock In \emph{Proceedings of the 2023 Conference on Empirical Methods in Natural Language Processing, {EMNLP} 2023}, pages 7101--7125.

\bibitem[{Zeng et~al.(2023)Zeng, Wei, Liu, and Fu}]{Zero-CQG}
Hongwei Zeng, Bifan Wei, Jun Liu, and Weiping Fu. 2023.
\newblock Synthesize, prompt and transfer: Zero-shot conversational question generation with pre-trained language model.
\newblock In \emph{Proceedings of the 61st Annual Meeting of the Association for Computational Linguistics, {ACL} 2023}, pages 8989--9010.

\bibitem[{Zhang et~al.(2023)Zhang, Chen, Shen, Ding, Tenenbaum, and Gan}]{mcts2}
Shun Zhang, Zhenfang Chen, Yikang Shen, Mingyu Ding, Joshua~B. Tenenbaum, and Chuang Gan. 2023.
\newblock Planning with large language models for code generation.
\newblock In \emph{Proceedings of the Eleventh International Conference on Learning Representations, {ICLR} 2023}, pages 1--28.

\bibitem[{Zhao et~al.(2023)Zhao, Lee, and Hsu}]{mcts3}
Zirui Zhao, Wee~Sun Lee, and David Hsu. 2023.
\newblock Large language models as commonsense knowledge for large-scale task planning.
\newblock In \emph{Proceedings of 37th Conference on Neural Information Processing Systems, {NeurIPS} 2023}, pages 1--21.

\end{thebibliography}
\appendix

\section{Experiment}
\subsection{Experimental Implementation Details}
\label{sec:append_exp}

\vpara{Parameters.} We configure the number of possible actions $k$ for each node to be 5, set the number of simulations to 10, and assign the exploration weight $w$ to 1. For both open-source and closed-source LLMs used, we set $n$ to 1, the $temperature$ to 0.7, and $top\_p$ to 1.

\begin{figure}[!t]
\centering 
\includegraphics[width=0.45\textwidth]{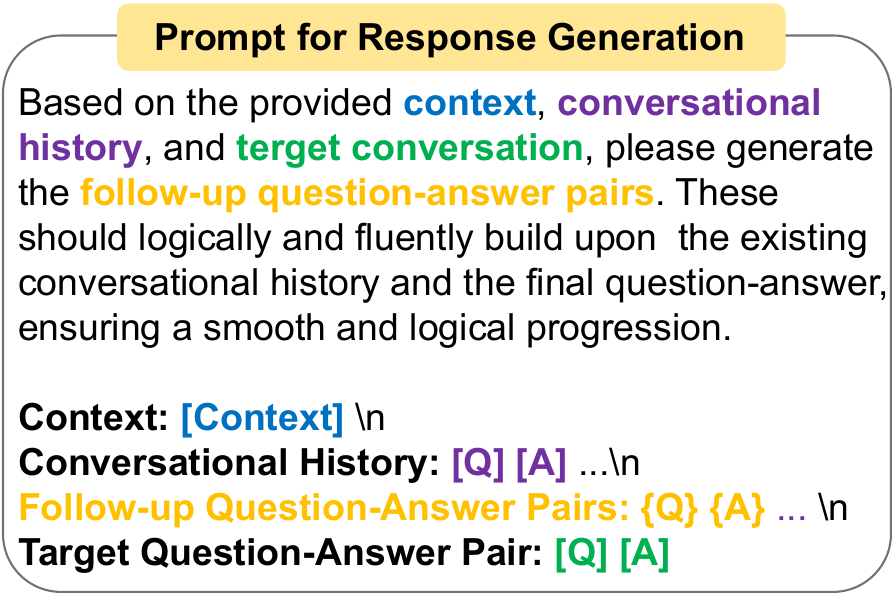}
\caption{The prompt for response generation.}
\label{fig:prompt_res} 
\end{figure}

\vpara{Prompt for Response Generation.} To guide LLMs in generation, we provide prompts. 
Instead of meticulously designing these, we focus on ensuring they effectively convey the intended meaning.
As illustrated in Figure~\ref{fig:prompt_res}, we present the prompt used in this work for generating the response.

\begin{figure}[!t]
\centering 
\includegraphics[width=0.45\textwidth]{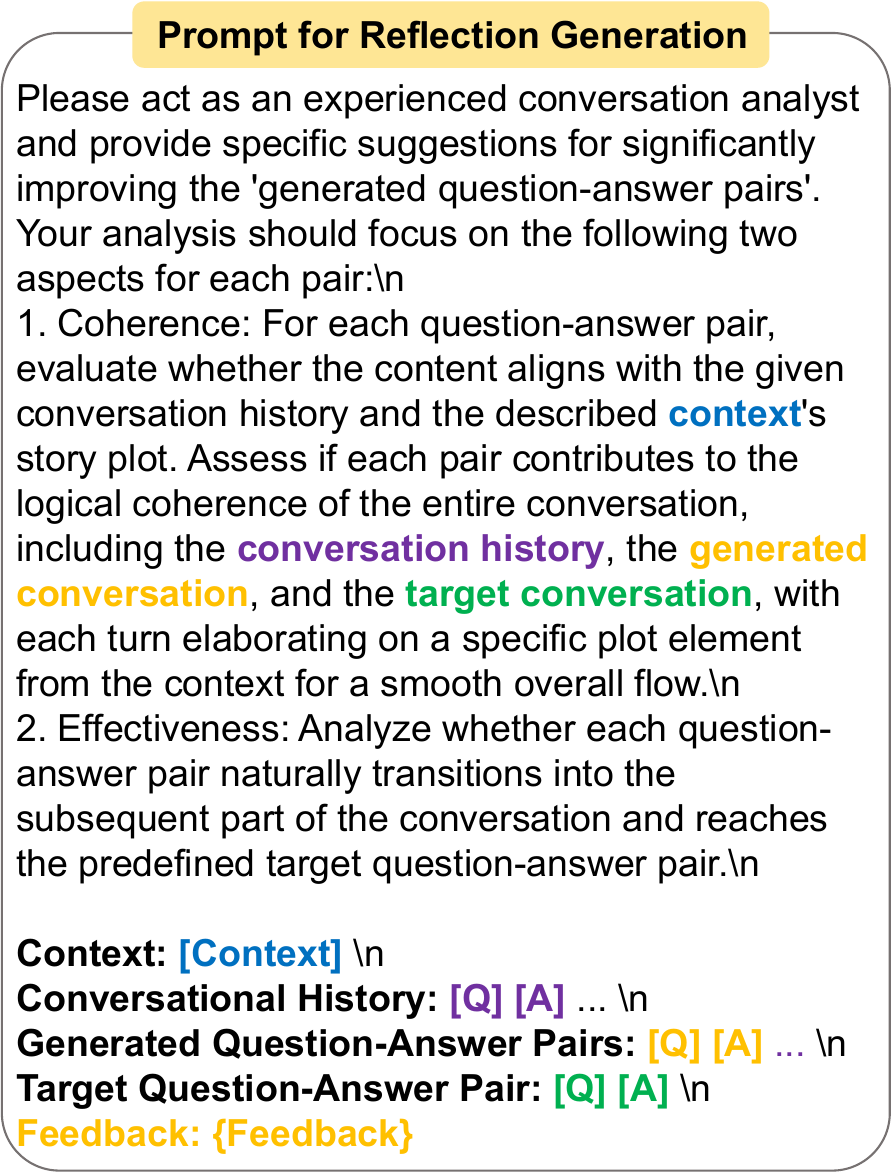}
\caption{The prompt for reflection generation.}
\label{fig:prompt_ref} 
\end{figure}

\vpara{Prompt for Reflection Generation.} Reflection aims to provide feedback on the entire planning path, enabling continuous self-assessment and adjustments to achieve the specified outcome successfully. Similar to the prompt for generating responses, we provide them casually rather than meticulously crafting them.
Figure~\ref{fig:prompt_ref} illustrates the prompt we utilized to generate the reflection.

\begin{figure}[!t]
\centering 
\includegraphics[width=0.45\textwidth]{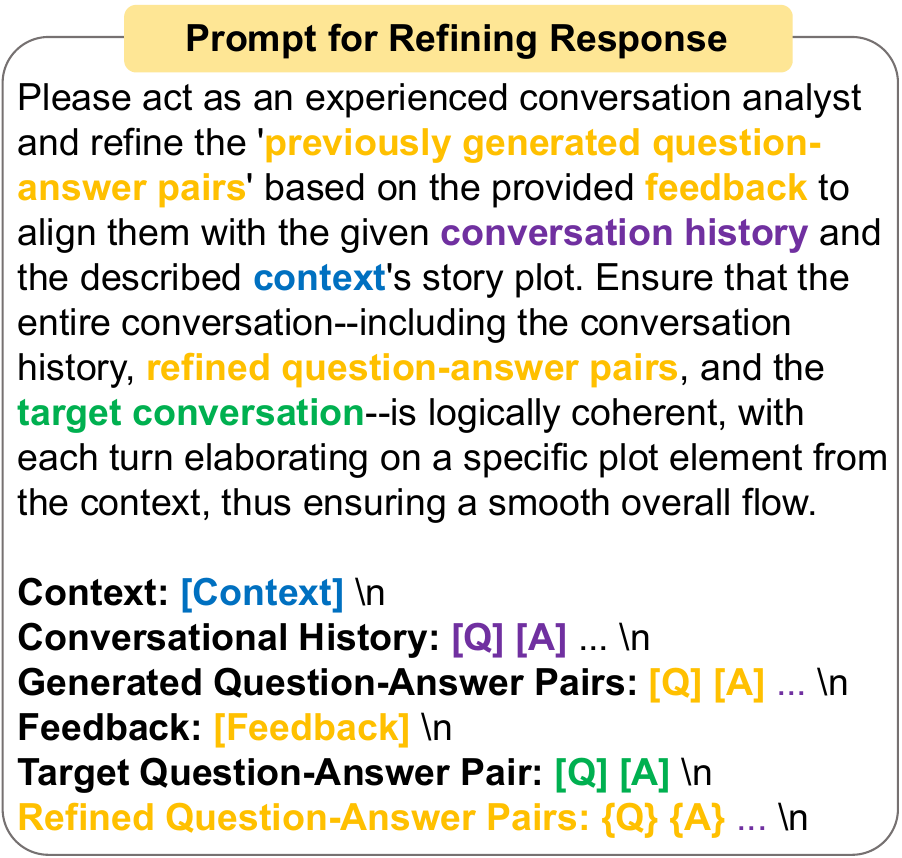}
\caption{The prompt for refining response generation involves obtaining detailed feedback from LLMs.}
\label{fig:prompt_refine} 
\end{figure}

\vpara{Prompt for Refining Response via Feedback.} Our proposed reflection mechanism produces insightful feedback for each action along the entire planning path. 
This feedback is subsequently used to formulate prompts for the LLMs, thereby enhancing their responses.
Similarly, we casually provide the prompts rather than crafting them meticulously.
As shown in Figure~\ref{fig:prompt_refine}, we present the prompt for refining response generation.

\subsection{Human Evaluation Scoring Guidelines}
\label{sec:append_human}

\begin{figure}[!t]
\centering 
\includegraphics[width=0.45\textwidth]{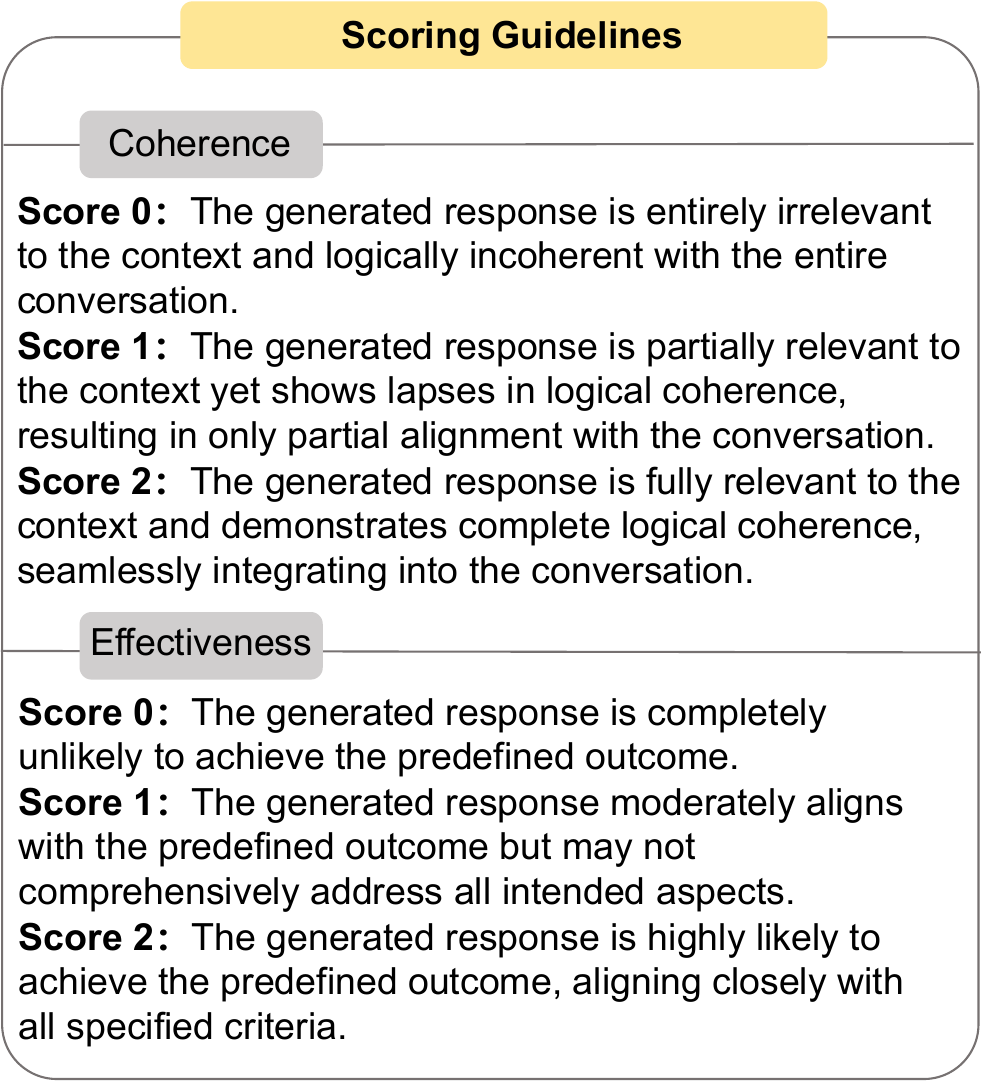}
\caption{Human evaluation scoring guidelines for two criteria: Coherence and Effectiveness.}
\label{fig:human_score} 
\end{figure}

We detail our scoring method used to guide three annotators in evaluating the generated responses based on two key criteria: Coherence and Effectiveness. Each criterion is scored on a scale of 0, 1, or 2, with higher scores indicating better performance. These criteria are explained in more detail in Section~\ref{sec:exp_settings} and illustrated in Figure~\ref{fig:human_score}.
Coherence assesses the logical flow and relevance of responses, whereas Effectiveness measures their success in achieving the specified outcome. This methodical approach ensures a consistent and unbiased assessment of response quality.

\begin{figure}[!t]
\centering 
\includegraphics[width=0.45\textwidth]{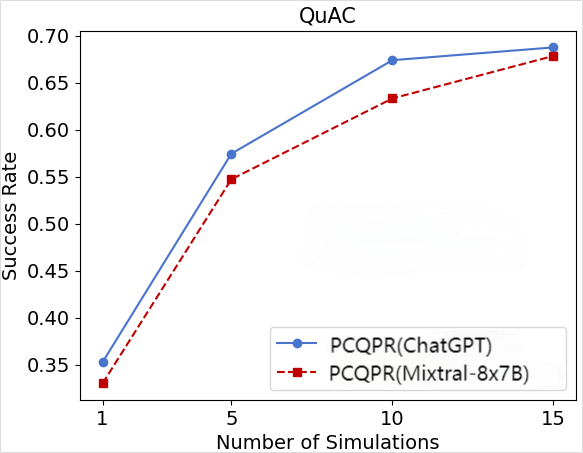}
\caption{Sensitivity study.}
\label{fig:sensitivity} 
\end{figure}

\subsection{Sensitivity Study}

We investigate the effect of varying the number of testing simulations on the performance of PCQPR. As illustrated in Figure~\ref{fig:sensitivity}, the Success Rate of PCQPR on QuAC varies with the number of simulations, showing an initial increase in performance followed by a more gradual and steady increase upon reaching 10 simulations.
In our experiments, we set the number of simulations to 10 to balance the performance gains against the associated costs. Beyond this threshold, while performance may continue to improve, the rate of increase slows significantly, indicating diminishing returns. Therefore, setting the number of simulations to 10 represents an optimal compromise between achieving substantial performance improvements and maintaining reasonable cost efficiency.

\subsection{Case Studies}
\label{sec:append_cs}

\begin{figure*}[!t]
\centering 
\includegraphics[width=1.0\textwidth]{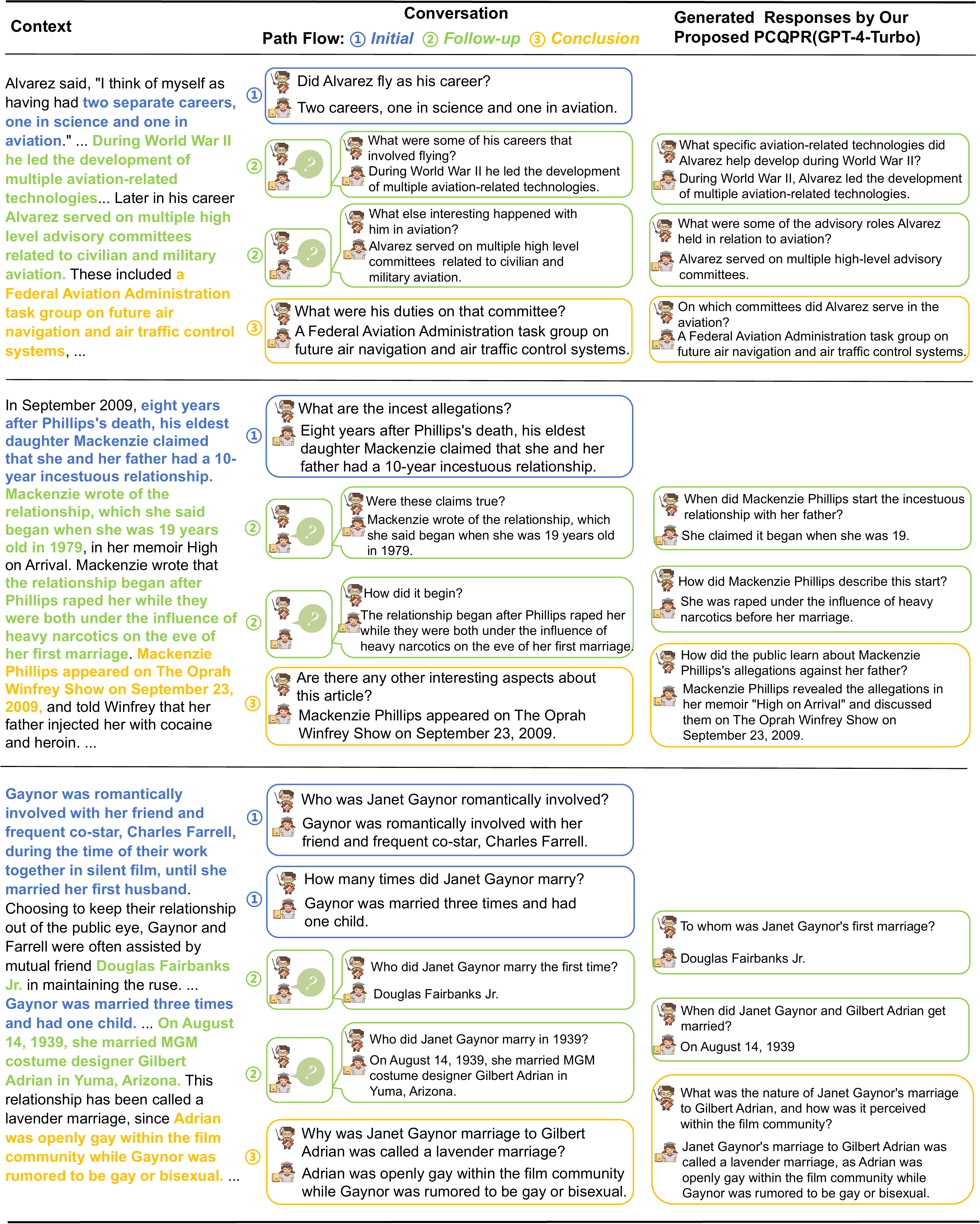}
\caption{Case studies for three examples generated by our method, PCQPR(GPT-4-Turbo).}
\label{fig:case_study} 
\end{figure*}

To visualize our model's performance, Figure~\ref{fig:case_study} presents three examples generated by PCQPR (GPT-4-Turbo). These examples demonstrate the model's ability to proactively steer the generation process to achieve the specified outcome. This showcases the model's advanced understanding and manipulation of context, ensuring that each QA pair is not only relevant but also precisely aligned with the desired outcome.

\end{document}